\documentclass[a4paper,a4wide]{article}
\usepackage{aaai}
\usepackage{times}
\usepackage{helvet}
\usepackage{courier}

\nocopyright

\usepackage[utf8]{inputenc}
\usepackage{natbib}
\usepackage[british]{babel}
\usepackage{relsize}
\usepackage{stmaryrd}
\usepackage{url}
\usepackage[font=it]{caption}
\usepackage{tikz,pgf}
\usetikzlibrary{shapes,backgrounds}
\usepackage{amsmath,amssymb}
\usepackage[amsmath,thref,thmmarks]{ntheorem}
\usepackage[pdftex,colorlinks=true,linkcolor=black,citecolor=black,menucolor=black,urlcolor=black]{hyperref}
\usepackage{cleveref}\renewcommand{\thref}[1]{\Cref{#1}} 
\usepackage{multirow}
\usepackage{comment}

\theoremstyle{plain}
\theoremseparator{.}
\theoremheaderfont{\normalfont\bfseries}
\theorembodyfont{\slshape}
\newtheorem{theorem}{Theorem}
\newtheorem{lemma}[theorem]{Lemma}
\newtheorem{proposition}[theorem]{Proposition}
\newtheorem{corollary}[theorem]{Corollary}
\theorembodyfont{\upshape}
\newtheorem{definition}{Definition}
\newtheorem{example}{Example}
\theoremstyle{nonumberplain}
\theoremheaderfont{\normalfont\itshape}
\newtheorem{xproof}{Proof}
\theoremstyle{nonumberplain}
\theoremheaderfont{\normalfont\itshape}
\theoremsymbol{\ensuremath{\Box}}
\newtheorem{proof}{Proof}

\ifdefined\implies
 \renewcommand{\implies}{\rightarrow}
\else
\newcommand{\implies}{\rightarrow}
\fi
\newcommand{\limplies}{\rightarrow}
\newcommand{\liff}{\leftrightarrow}
\newcommand{\lniff}{\nleftrightarrow}

\newcommand{\define}[1]{\emph{#1}}
\newcommand{\set}[1]{\left\{#1\right\}}
\newcommand{\guard}{\ \middle\vert\ }
\newcommand{\eqdef}{\,\stackrel{\raisebox{-.2ex}{\tiny\normalfont def}}{\raisebox{-.3ex}{$=$}}\,}
\renewcommand{\eqdef}{=}

\newcommand{\cardinality}[1]{\left\lvert#1\right\rvert}

\newcommand{\cc}[1]{{\textup{\textsf{#1}}}}

\newcommand{\NP}{\ensuremath{\cc{NP}}}

\newcommand{\SigmaP}[1]{\ensuremath{\Sigma_{#1}^{P}}}

\newcommand{\tvt}{\mathbf{t}}
\newcommand{\tvf}{\mathbf{f}}
\newcommand{\lin}{\tvt}

\newcommand{\expressive}{e}

\newcommand{\eleq}{\leq_\expressive}

\newcommand{\lt}{<}
\newcommand{\elt}{\lt_\expressive}

\newcommand{\eeq}{\cong_\expressive}

\newcommand{\su}{\mathit{su}}
\newcommand{\st}{\mathit{st}}
\newcommand{\model}{\mathit{mod}}
\newcommand{\supp}[1]{#1^\su}
\newcommand{\stab}[1]{#1^\st}

\newcommand{\AF}{\textrm{AF}}
\newcommand{\LP}{\textrm{LP}}
\newcommand{\BADF}{\textrm{BADF}}
\newcommand{\ADF}{\textrm{ADF}}
\newcommand{\PL}{\textrm{PL}}

\newcommand{\LPsu}{\supp{\LP}}
\newcommand{\LPst}{\stab{\LP}}
\newcommand{\BADFsu}{\supp{\BADF}}
\newcommand{\BADFst}{\stab{\BADF}}
\newcommand{\ADFsu}{\supp{\ADF}}
\newcommand{\ADFst}{\stab{\ADF}}

\newcommand{\bsup}{\mathit{sup}}
\newcommand{\batt}{\mathit{att}}

\newcommand{\parents}[1]{\mathit{par}(#1)}

\newcommand{\pin}[2]{p^{#1}_{#2}}
\newcommand{\psup}[2]{p_{\bsup}^{#1,#2}}
\newcommand{\patt}[2]{p_{\batt}^{#1,#2}}

\newcommand{\lp}{P}
\newcommand{\lpif}{\leftarrow}

\DeclareMathOperator{\lpnot}{\mathit{not}\,} 
\newcommand{\literals}[1]{{#1}^\pm}

\begin{document}

\title{%
  On the Relative Expressiveness of \mbox{Argumentation Frameworks}, \mbox{Normal Logic Programs} and \mbox{Abstract Dialectical Frameworks}
}
\author{%
Hannes Strass\\
Computer Science Institute\\
Leipzig University, Germany
}

\maketitle

\begin{abstract}
  We analyse the expressiveness of the two-valued semantics of 
  abstract argumentation frameworks,
  normal logic programs and 
  abstract dialectical frameworks.
  By expressiveness we mean the ability to encode a desired set of two-valued interpretations over a given propositional signature using only atoms from that signature.
  While the computational complexity of the two-valued model existence problem for all these languages is (almost) the same, we show that the languages form a neat hierarchy with respect to their expressiveness.
\end{abstract}


\section{Introduction}
\label{sec:introduction}

More often than not, different knowledge representation languages have conceptually similar and partially overlapping intended application areas.
What are we to do if faced with an application and a choice of several possible knowledge representation languages which could be used for the application?
One of the first axes along which to compare different formalisms that comes to mind is computational complexity:
if a language is computationally too expensive when considering the problem sizes typically encountered in practice, then this is a clear criterion for exclusion.

But what if the available language candidates have the same computational complexity?
If their expressiveness in the computational-complexity sense of ``What kinds of \emph{problems} can the formalism solve?'' is the same, we need a more fine-grained notion of expressiveness.
In this paper, we use such an alternative notion and perform an exemplary study of the relative expressiveness of several different knowledge representation languages:
argumentation frameworks (AFs)~\citep{dung95acceptability},
normal logic programs (LPs),
abstract dialectical frameworks (ADFs)~\citep{brewka-woltran10adfs}
and propositional logic.

This choice of languages is largely motivated by the similar intended application domains of argumentation frameworks and abstract dialectical frameworks and the close relation of the latter to normal logic programs.
We add propositional logic to have a well-known reference point.
Furthermore, the computational complexity of their respective model existence problems is the same (with one exception):
\begin{itemize}
\item for AFs, deciding stable extension existence is \NP-complete~\citep{dimopoulos02complexity};
\item for LPs, deciding the existence of supported/stable models is \NP-complete~\citep{bidoit91negation,marek91autoepistemic};
\item for ADFs, deciding the existence of models is \NP-complete~\citep{brewka13adfs}, deciding the existence of stable models is \SigmaP{2}-complete for general ADFs~\citep{brewka13adfs} and \NP-complete for the subclass of bipolar ADFs~\citep{strass-wallner14complexity};
\item the satisfiability problem of propositional logic is \NP-complete. 
\end{itemize}

In view of these almost identical complexities, we use an alternative measure of the expressiveness of a knowledge representation language $L$:
``Given a set of two-valued interpretations, is there a knowledge base in $L$ that has this exact model set?''
This notion lends itself straightforwardly to compare different formalisms~\citep{gogic95comparative}:
\begin{quote}
  Formalism $L_2$ is at least as expressive as formalism $L_1$ if and only if
  every knowledge base in $L_1$ has an equivalent knowledge base in $L_2$.
\end{quote}
So here expressiveness is understood in terms of \emph{realisability}, ``What kinds of model sets can the formalism express?''

It is easy to see that propositional logic can express any set of two-valued interpretations.
The same is easy (but less easy) to see for logic programs under supported model semantics.
For logic programs under stable model semantics, it is clear that not all model sets can be expressed, since two different stable models are always incomparable with respect to the subset relation.
In this paper, we study such expressiveness properties for all the mentioned formalisms under different semantics.
It will turn out that the languages form a more or less strict expressiveness hierarchy, with AFs at the bottom, ADFs and LPs under stable semantics higher up and ADFs and LPs under supported model semantics at the top together with propositional logic.

To show that a language $L_2$ is at least as expressive as a language $L_1$ we will mainly use two different techniques.
In the best case, we can use a syntactic compact and faithful translation from knowledge bases of $L_1$ to those of $L_2$.
\define{Compact} means that the translation does not change the vocabulary, that is, does not introduce new atoms.
\define{Faithful} means that the translation exactly preserves the models of the knowledge base for respective semantics of the two languages.
In the second best case, we assume given the knowledge base of $L_1$ in the form of a set $X$ of desired models and construct a semantic \define{realisation} of $X$ in $L_2$, that is, a knowledge base in $L_2$ whose model set corresponds exactly to $X$.
To show that language $L_2$ is \emph{strictly more expressive} than $L_1$, we additionally have to present a knowledge base $K$ from $L_2$ of which we prove that $L_1$ cannot express the model set of $K$.

For all methods, we can make use of several recent works on the formalisms we study here.
First of all, we~[\citeyear{strass13approximating}] studied the syntactic intertranslatability of ADFs and LPs, but did not look at expressiveness or realisability.
The latter was recently studied for argumentation frameworks by \citet{dunne14characteristics}.
They allow to extend the vocabulary in order to realise a given model set, as long as the new vocabulary elements are evaluated to false in all models.
For several semantics of AFs, \citeauthor{dunne14characteristics} found necessary (and sufficient) conditions for realisability.
While their sufficient conditions are not applicable to our setting, they discovered a necessary condition for realisability with stable extension semantics that we will make use of in this paper.
There has also been work on translating ADFs into AFs for the ADF model and AF stable extension semantics~\citep{brewka11relating}, however this translation introduces additional arguments and is therefore not compact.

The gain that is achieved by our results is not only that of increased clarity about fundamental properties of these knowledge representation languages -- \emph{What can these formalisms express, actually?} -- but has several further applications.
As \citet{dunne14characteristics} remarked, a major application is in constructing knowledge bases with the aim of encoding a certain model set.
As a necessary prerequisite to this, it must be known that the intended model set is realisable in the first place.
For example, in a recent approach to revising argumentation frameworks~\citep{coste13revision}, the authors avoid this problem by assuming to produce a \emph{collection} of AFs whose model sets in union produce the desired model set.
While the work of \citet{dunne14characteristics} showed that this is indeed necessary in the case of AFs and stable extension semantics (that is, there are model sets that a single AF just cannot express), our work shows that for ADFs under the model semantics, a single knowledge base (ADF) is always enough to realise any given model set.

Of course, the fact that the languages we study have the same computational complexity means that there in principle exist polynomial intertranslations for the respective decision problems.
But such intertranslations may involve the introduction of new atoms.
In theory, a polynomial blowup from $n$ atoms to $n^k$ atoms for some $k$ is of no consequence.
In practice, it has a profound impact:
the number $n$ of atoms directly influences the search space that any implementation potentially has to cover.
There, an increase from $2^n$ to \mbox{$2^{n^k}$} is no longer polynomial, but exponential, and accordingly makes itself felt.
Being able to realise a model set compactly, without new atoms, therefore attests that a language $L$ has a certain basic kind of efficiency property, in the sense that the $L$-realisation of a model set does not unnecessarily enlarge the search space of algorithms operating on it.

The paper proceeds as follows.
We first define the notion of expressiveness formally and then introduce the languages we will study.
After reviewing several intertranslatability results for these languages, we stepwise obtain the results that lead to the expressiveness hierarchy.
We conclude with a discussion of avenues for future work.

\section{Background}
\label{sec:background}

We assume given a finite set \mbox{$A$} of atoms (statements, arguments), the \define{vocabulary}.
A knowledge representation language interpreted over $A$ is then some set $L$;
a (two-valued) semantics for $L$ is a mapping \mbox{$\sigma : L \to 2^{2^A}$} that assigns sets of two-valued models to the language elements.
(So $A$ is implicit in $L$.)
Strictly speaking, a two-valued interpretation is a mapping from the set of atoms into the two truth values true and false, but for technical ease we represent two-valued interpretations by the sets containing the atoms that are 
true.

For a language $L$, we denote the range of the semantics $\sigma$ by $\sigma(L)$.
Intuitively, $\sigma(L)$ is the set of models that language $L$ can express, with any knowledge base over vocabulary $A$ whatsoever.
For example, for \mbox{$L=\PL$} propositional logic and \mbox{$\sigma=\model$} the usual model semantics, we have \mbox{$\sigma(\PL)=2^{2^A}$} since obviously any set of models is realisable in propositional logic.%
\footnote{For a set \mbox{$X\subseteq 2^A$} we can simply define \mbox{$\varphi_X = \bigvee_{M\in X} \varphi_M$} with \mbox{$\varphi_M = \bigwedge_{a\in M}a \land \bigwedge_{a\in A\setminus M}\neg a$} and clearly \mbox{$\model(\varphi_X)=X$}.}
This leads us to compare different pairs of languages and semantics with respect to the semantics' range of models.
Our concept of ``language'' concentrates on semantics and decidedly remains abstract.
\begin{definition}
  \label{def:eleq}
  Let $A$ be a finite vocabulary, 
  $L_1,L_2$ be languages that are interpreted over $A$ and
  \mbox{$\sigma_1:L_1\to 2^{2^A}$} and \mbox{$\sigma_2:L_2\to 2^{2^A}$} be two-valued semantics.
  We define
  \begin{gather*}
    L_1^{\sigma_1} \eleq L_2^{\sigma_2}
    \quad\text{iff}\quad
    \sigma_1(L_1) \subseteq \sigma_2(L_2)
  \end{gather*}
\end{definition}
Intuitively, language $L_2$ under semantics $\sigma_2$ is at least as expressive as language $L_1$ under semantics $\sigma_1$, because all models that $L_1$ can express under $\sigma_1$ are also contained in those that $L_2$ can produce under $\sigma_2$.
(If the semantics are clear from the context we will omit them;
this holds in particular for argumentation frameworks and propositional logic, where we only look at a single semantics.)
As usual, 
\begin{itemize}
\item \mbox{$L_1 \elt L_2$} iff \mbox{$L_1 \eleq L_2$} and \mbox{$L_2 \not\eleq L_1$};
\item \mbox{$L_1 \eeq L_2$} iff \mbox{$L_1 \eleq L_2$} and \mbox{$L_2 \eleq L_1$}.
\end{itemize}
The relation $\eleq$ is reflexive and transitive by definition, but not necessarily antisymmetric.
That is, there might different languages \mbox{$L_1\neq L_2$} that are equally expressive: \mbox{$L_1\eeq L_2$}.

We next introduce the particular knowledge representation languages we study in this paper.
All will make use of a vocabulary $A$; the results of the paper are all considered parametric in such a given vocabulary.

\subsection{Logic Programs}
\label{sec:logic-programs}

For a vocabulary $A$ define \mbox{$\lpnot A = \set{ \lpnot a \guard a \in A }$} and the set of literals over $A$ as \mbox{$\literals{A} = A \cup \lpnot A$}.
A \define{normal logic program rule} over $A$ is then of the form \mbox{$a \lpif B$} where $a \in A$ and $B \subseteq \literals{A}$.
The rule can be read as logical consequence, ``$a$ is true if all literals in $B$ are true.''
The set $B$ is called the \define{body} of the rule, we denote by \mbox{$B^+ = B \cap A$} and \mbox{$B^- = \set{ a \in A \guard \lpnot a \in B }$} the \define{positive} and \define{negative body} atoms, respectively.
A rule is \define{definite} if \mbox{$B^- = \emptyset$}.
For singleton \mbox{$B = \set{ b }$} we denote the rule just by \mbox{$a \lpif b$}.
A \define{logic program (LP)} $\lp$ over $A$ is a set of logic program rules over $A$, and it is definite if all rules in it are definite.

At first, logic programs were restricted to definite programs, whose semantics was defined through the proof-theoretic procedure of SLD resolution.
The meaning of negation $\lpnot$ was only defined operationally through negation as failure.
\citet{clark78negation} gave the first declarative semantics for normal logic programs via a translation to classical logic that will be recalled shortly.
This leads to the {supported model semantics} for logic programs:
A rule \mbox{$a\lpif B\in \lp$} is \define{active} in a set \mbox{$M\subseteq A$} iff
\mbox{$B^+\subseteq M$} and \mbox{$B^-\cap M=\emptyset$} imply \mbox{$a\in M$}.
$M$ is a \define{supported model} for $\lp$ iff \mbox{$M=\set{ a\in A \guard a\lpif B\in \lp \text{ is active in } M }$}.
For a logic program $\lp$ we denote the set of its supported models by $\su(\lp)$.
The intuition behind this semantics is that everything that is true in a model has some kind of support.

However, this support might be cyclic self-support.
For instance, the logic program \mbox{$\set{a\lpif a}$} has two supported models, $\emptyset$ and $\set{a}$, where the latter is undesired in many application domains.
As an alternative, \citet{gelfond-lifschitz88thestablemodel} proposed the stable model semantics, a declarative semantics for negation as failure that does not allow self-support:
\mbox{$M\subseteq A$} is a \define{stable model} for $\lp$ iff $M$ is the $\subseteq$-least supported model of ${\lp^M}$, where the definite program $\lp^M$ is obtained from $\lp$ by
(1) eliminating each rule whose body contains a literal \mbox{$\lpnot a$} with \mbox{$a \in M$}, and
(2) deleting all literals of the form \mbox{$\lpnot a$} from the bodies of the remaining rules.
We write $\st(\lp)$ for the set of stable models of $\lp$.
It follows from the definition of stable models that $\st(\lp)$ is a $\subseteq$-antichain: for all \mbox{$M_1\neq M_2\in \st(\lp)$} we have \mbox{$M_1\not\subseteq M_2$}.



\subsection{Argumentation Frameworks}
\label{sec:argumentation-frameworks}

\citet{dung95acceptability} introduced argumentation frameworks as pairs \mbox{$F=(A,R)$} where $A$ is a set and \mbox{$R \subseteq A \times A$} a relation.
The intended reading of an AF $F$ is that the elements of $A$ are arguments whose internal structure is abstracted away.
The only information about the arguments is given by the relation $R$ encoding a notion of attack: 
a pair \mbox{$(a,b) \in R$} expresses that argument $a$ attacks argument $b$ in some sense.

The purpose of semantics for argumentation frameworks is to determine sets of arguments (called \define{extensions}) which are acceptable according to various standards.
For a given extension \mbox{$S\subseteq A$}, the arguments in $S$ are considered to be accepted, those that are attacked by some argument in $S$ are considered to be rejected, and all others are neither, their status is undecided.
We will only be interested in so-called \define{stable} extensions, sets $S$ of arguments that do not attack each other and attack all arguments not in the set.
For stable extensions, each argument is either accepted or rejected by definition, thus the semantics is two-valued.
More formally, a set \mbox{$S \subseteq A$} of arguments is \define{conflict-free} iff there are no \mbox{$a,b \in S$} with \mbox{$(a,b) \in R$}.
A set $S$ is a \define{stable extension} for \mbox{$(A,R)$} iff it is conflict-free and for all \mbox{$a\in A\setminus S$} there is a \mbox{$b\in S$} with \mbox{$(b,a)\in R$}.
For an AF $F$, we denote the set of its stable extensions by $\st(F)$.
Again, it follows from the definition of a stable extension that the set $\st(F)$ is always a $\subseteq$-antichain.

\subsection{Abstract Dialectical Frameworks}
\label{sec:abstract-dialectical-frameworks}

An abstract dialectical framework (ADF) is a directed graph whose
nodes represent statements or positions which can be accepted or
not. The links represent dependencies: the status of a node $a$ only
depends on the status of its parents (denoted $\parents{a}$), that
is, the nodes with a direct link to $a$. In addition, each node $a$
has an associated acceptance condition $C_a$ specifying the exact
conditions under which $a$ is accepted. $C_a$ is a function
assigning to each subset of $\parents{a}$ one of the truth values $\tvt$ or
$\tvf$. Intuitively, if for some \mbox{$R \subseteq \parents{a}$} we have 
\mbox{$C_a(R) = \tvt$}, then $a$ will be accepted provided the nodes in $R$ are accepted and those in \mbox{$\parents{a}\setminus R$} are not accepted.

More formally, an \emph{abstract dialectical framework} is a tuple \mbox{$D = (A, L, C)$} where
\begin{itemize}
\item $A$ is a set of statements,
\item $L \subseteq A \times A$ is a set of links,
\item \mbox{$C = \set{C_a}_{a \in A}$} is a collection of total functions 
  \mbox{$C_a : 2^{\parents{a}}\to \set{\tvt,\tvf}$}, one for each statement $a$. 
  The function $C_a$ is called \define{acceptance condition of $a$}.
\end{itemize}
It is often convenient to represent acceptance conditions by propositional formulas.
In particular, we will do so for several results of this paper.
There, each $C_a$ is represented by a propositional formula \mbox{$\varphi_a$} over $\parents{a}$.
Then, clearly, \mbox{$C_a(R\cap\parents{a})=\lin$} iff $R$ is a model for $\varphi_a$, \mbox{$R\models\varphi_a$}.

\citet{brewka-woltran10adfs} introduced a useful subclass of ADFs:
an ADF \mbox{$D = (A, L, C)$} is \define{bipolar} iff all links in $L$ are supporting or attacking (or both).
A link \mbox{$(b,a) \in L$} is \define{supporting in $D$} iff for all \mbox{$R \subseteq \parents{a}$}, we have that \mbox{$C_a(R)=\tvt$} implies \mbox{$C_a(R \cup \set{b})=\tvt$}.
Symmetrically, a link \mbox{$(b,a) \in L$} is \define{attacking in $D$} iff for all \mbox{$R \subseteq \parents{a}$}, we have that \mbox{$C_a(R \cup \set{b})=\tvt$} implies \mbox{$C_a(R)=\tvt$}.
If a link \mbox{$(b,a)$} is both supporting and attacking then $b$ has no influence on $a$, the link is redundant (but does not violate bipolarity).
We will sometimes use this circumstance when searching for ADFs;
there we simply assume that \mbox{$L=A \times A$}, then links that are actually not needed can be expressed by acceptance conditions that make them redundant.

There are numerous semantics for ADFs;
we will only be interested in two of them, (supported) models and stable models.
A set \mbox{$M\subseteq A$} is a \define{model of $D$} iff for all \mbox{$a\in A$} we find that \mbox{$a\in M$} iff \mbox{$C_a(M)=\tvt$}.
The definition of stable models is inspired by logic programming and slightly more complicated~\citep{brewka13adfs}.
Define an operator by \mbox{$\Gamma_D(Q, R) = (\mathit{acc}(Q, R), \mathit{rej}(Q, R))$} for \mbox{$Q,R\subseteq A$}, where
\def\negspace{\hspace*{-6mm}}
\begin{align*}
  \mathit{acc}(Q, R) &= \{ a \in A \mid &&\negspace\text{for all } Q \subseteq Z \subseteq (A\setminus R), \\
  &&&\negspace\text{we have } C_a(Z)=\tvt \} \\
  \mathit{rej}(Q, R) &= \{ a \in A \mid &&\negspace\text{for all } Q \subseteq Z \subseteq (A\setminus R), \\
  &&&\negspace\text{we have } C_a(Z)=\tvf \}
\end{align*}
The intuition behind the operator is as follows:
A pair $(Q,R)$ represents a partial interpretation of the set of statements where those in $Q$ are accepted (true), those in $R$ are rejected (false), and those in $S\setminus (Q\cup R)$ are neither.
The operator checks for each statement $a$ whether all total interpretations that can possibly arise from $(Q,R)$ agree on their truth value for the acceptance condition for $a$.
That is, if $a$ has to be accepted no matter how the statements in \mbox{$S\setminus (Q\cup R)$} are interpreted, then $a\in\mathit{acc}(Q,R)$.
The set $\mathit{rej}(Q,R)$ is computed symmetrically, so the pair \mbox{$(\mathit{acc}(Q, R), \mathit{rej}(Q, R))$} constitutes a refinement of $(Q,R)$.

For \mbox{$M \subseteq A$}, the reduced ADF \mbox{$D^M = (M, L^M, C^M)$} is defined by 
\mbox{$L^M=L\cap M\times M$} and 
for each \mbox{$a\in M$} setting \mbox{$\varphi_a^M = \varphi_a[b/\tvf : b\notin M]$}, that is, replacing all \mbox{$b\notin M$} by false in the acceptance formula of $a$.
A model $M$ for $D$ is a \define{stable model} of $D$ iff
the least fixpoint of the operator $\Gamma_{D^M}$ is given by $(M, \emptyset)$.
As usual, $\su(D)$ and $\st(D)$ denote the model sets of the two semantics.
While ADF models can be subsets of one another, ADF stable models cannot.


\subsection{Translations between the formalisms}
\label{sec:translations}

\subsubsection*{From AFs to BADFs} 
\citet{brewka-woltran10adfs} showed how to translate AFs into ADFs:
For an AF \mbox{$F=(A, R)$}, define the ADF associated to $F$ as 
\mbox{$D(F)=(A,R,C)$} with \mbox{$C=\set{\varphi_a}_{a\in A}$} and
\mbox{$\varphi_a=\bigwedge_{(b,a)\in R}\neg b$} for $a\in A$.
Clearly, the resulting ADF is bipolar; parents are always attacking.
\citet{brewka-woltran10adfs} proved that this translation is faithful for the AF stable extension and ADF model semantics (Proposition~1).
\citet{brewka13adfs} later proved the same for the AF stable extension and ADF stable model semantics (Theorem~4).
It is easy to see that the translation can be computed in polynomial time.

\subsubsection*{From ADFs to PL}
\citet{brewka-woltran10adfs} also showed that ADFs under supported model semantics can be faithfully translated into propositional logic:
When acceptance conditions of statements \mbox{$a\in A$} are represented by propositional formulas $\varphi_a$, then the supported models of an ADF $D$ over $A$ are given by the classical models of the formula set \mbox{$\set{ a\liff\varphi_a \guard a\in A }$}.

\subsubsection*{From AFs to PL}
In combination, the previous two translations yield a polynomial and faithful translation chain from AFs into propositional logic.

\subsubsection*{From ADFs to LPs} 
In recent work we showed that ADFs can be faithfully translated into normal logic programs~\citep{strass13approximating}.
For an ADF \mbox{$D = (A,L,C)$}, its standard logic program $\lp(D)$ is given by
\begin{gather*}
  \set{ a \lpif (M \cup \lpnot (\parents{a} \setminus M)) \guard a \in A, C_a(M)=\tvt }
\end{gather*}
It is an easy consequence of Lemma~3.14 in \citep{strass13approximating} that this translation preserves the supported model semantics.
For complexity reasons, we cannot expect that this translation is also faithful for the stable semantics.
And indeed, the ADF $D=(\set{a},\set{(a,a)},\set{\varphi_a=a\lor\neg a})$ has a stable model $\set{a}$ while its standard logic program $\lp(D) = \set{ a\lpif a, a\lpif \lpnot a }$ has no stable model.

\subsubsection*{From AFs to LPs}
The translation chain from AFs to ADFs to LPs is compact, and faithful for AF stable semantics and LP stable semantics~\citep{osorio05inferring}, and AF stable semantics and LP supported semantics~\citep{strass13approximating}.

\subsubsection*{From LPs to PL}
It is well-known that normal logic programs under supported model semantics can be translated to propositional logic~\citep{clark78negation}.
There, a logic program $\lp$ is translated to a propositional theory \mbox{$\Phi_\lp=\set{ a\liff\varphi_a \guard a\in A }$} where
\begin{gather*}
  \varphi_a = \bigvee_{a \lpif B \in \lp}\left(\bigwedge_{b \in B^+}b \wedge \bigwedge_{b \in B^-}\neg b\right)
\end{gather*}
for \mbox{$a \in A$}.
For the stable model semantics, additional formulas have to be added, but the extended translation works all the same~\citep{lin-zhao04assat}.

\subsubsection*{From LPs to ADFs}
The Clark completion of a normal logic program directly yields an equivalent ADF over the same signature~\citep{brewka-woltran10adfs}.
Clearly the translation is computable in polynomial time and the blowup (with respect to the original logic program) is at most linear.
The resulting translation is faithful for the supported model semantics, which is a straightforward consequence of Lemma~3.16 in \citep{strass13approximating}.

\section{Relative Expressiveness}
\label{sec:expressiveness}

We now analyse and compare the relative expressiveness of argumentation frameworks -- AFs --, (bipolar) abstract dialectical frameworks -- (B)ADFs --, normal logic programs -- LPs -- and propositional logic -- PL.
We first look at the different families of semantics -- supported and stable models -- in isolation and afterwards combine the two.
For the languages \mbox{$L\in\set{\ADF,\LP}$} that have both supported and stable semantics, we will indicate the semantics $\sigma$ via a superscript as in \Cref{def:eleq}.
For AFs we only consider the stable extension semantics, as this is (to date) the only two-valued semantics for AFs. 
For propositional logic PL we consider the usual model semantics.

With the syntactic translations we reviewed in the previous section, we currently have the following relationships.
For the supported semantics,
\begin{quote}
  \mbox{$\AF \eleq \BADFsu \eleq \ADFsu \eeq \LPsu \eleq \PL$}
\end{quote}
and for the stable semantics,
\begin{quote}
  \mbox{$\AF \eleq \BADFst \eleq \ADFst \elt \PL$} \\[3pt]
  \mbox{$\AF \eleq \LPst \elt \PL$}
\end{quote}
Note that \mbox{$\ADFst\elt\PL$} and \mbox{$\LPst\elt\PL$} hold since sets of stable models have an antichain property, in contrast to model sets of propositional logic.

\subsection{Supported semantics}
\label{sec:expressiveness:supported}

As depicted above, we know that expressiveness from AFs to propositional logic does not decrease.
However, it is not yet clear if any of the relationships is strict.

We first show that ADFs can realise any set of models.
To show this, we first make a case distinction whether the desired-model set is empty.
If there should be no model, we construct an ADF without models.
If the set of desired models is nonempty, we construct acceptance conditions directly from the set of desired interpretations.
The construction is similar in design to the one we reviewed for propositional logic, but takes into account the additional interaction between statements and their acceptance conditions.

\begin{theorem}
  \label{thm:realise:model}
  $\PL \eleq \ADFsu$
  \begin{proof}
    Consider a vocabulary $A$ and a set \mbox{$X\subseteq 2^A$}.
    We construct an ADF $D^\su_X$ with \mbox{$\su(D^\su_X)=X$} as follows.
    \begin{enumerate}
    \item 
      \mbox{$X=\emptyset$}.
      We choose some \mbox{$a\in A$} and set
      \mbox{$D^\su_X = (\set{a}, \set{(a,a)}, \set{C_a})$} with
      \mbox{$C_a(\emptyset) = \tvt$} and \mbox{$C_a(\set{a}) = \tvf$}.
      It is easy to see that $D^\su_X$ has no model.
    \item \mbox{$X\neq\emptyset$}.
      Define \mbox{$D^\su_X = (A, L, C)$} where \mbox{$L = A\times A$} and for each \mbox{$a\in A$} and \mbox{$M\subseteq A$}, we set \mbox{$C_a(M) = \tvt$}  iff 
      \begin{gather*}
        (M\in X \text{ and } a\in M) \text{ or }
        (M\notin X \text{ and } a\notin M )
      \end{gather*}
      We have to show that \mbox{$M\in X$} iff \mbox{$M \text{ is a model for } D^\su_X$}.
      \begin{description}
      \item[\normalfont ``if'':] 
        Let $M$ be a model of $D^\su_X$.
        \begin{enumerate}
        \item \mbox{$M=\emptyset$}.
          Pick any \mbox{$a\in A$}.
          Since $M$ is a model of $D^\su_X$, we have \mbox{$C_a(M)=\tvf$}.
          So either 
          (A) \mbox{$M\in X$} and \mbox{$a\notin M$} or 
          (B) \mbox{$M\notin X$} and \mbox{$a\in M$}, by definition of $C_a$.
          By assumption \mbox{$M=\emptyset$}, thus \mbox{$a\notin M$} and \mbox{$M\in X$}.
        \item \mbox{$M\neq\emptyset$}.
          Let \mbox{$a\in M$}.
          Then \mbox{$C_a(M)=\tvt$} since $M$ is a model of $D^\su_X$.
          By definition of $C_a$, \mbox{$M\in X$}.
        \end{enumerate}
      \item[\normalfont ``only if'':] Let $M\in X$.
        \begin{enumerate}
        \item \mbox{$M=\emptyset$}.
          Choose any \mbox{$a\in A$}.
          By assumption, \mbox{$a\notin M$} and \mbox{$M\in X$}, whence \mbox{$C_a(M)=\tvf$} by definition.
          Since \mbox{$a\in A$} was chosen arbitrarily, we have \mbox{$C_a(M)=\tvf$} iff \mbox{$a\notin M$}.
          Thus $M$ is a model of $D^\su_X$.
        \item \mbox{$M\neq\emptyset$}.
          Let \mbox{$a\in A$}.
          If \mbox{$a\in M$}, then by assumption and definition of $C_a$ we have \mbox{$C_a(M)=\tvt$}.
          Conversely, if \mbox{$a\notin M$}, then by definition \mbox{$C_a(M)=\tvf$}.
          Since \mbox{$a\in A$} was arbitrary, $M$ is a model of $D^\su_X$.
        \end{enumerate}
      \end{description}
    \end{enumerate}
  \end{proof}
\end{theorem}

When the acceptance conditions are written as propositional formulas, the construction in \thref{thm:realise:model} simply sets
\begin{align*}
  \varphi_a &\eqdef \bigvee_{M\in X,a\in M}\varphi_M \vee \bigvee_{M\subseteq A, M\notin X, a\notin M}\varphi_M \\
  \varphi_M &\eqdef \bigwedge_{a\in M}a \wedge \bigwedge_{a\in A\setminus M}\neg a 
\end{align*}

Since ADFs under supported semantics can be faithfully translated into logic programs, which can be likewise further translated to propositional logic, we have the following.

\begin{corollary}
  $\ADFsu \eeq \LPsu \eeq \PL$
\end{corollary}

While general ADFs under the supported model semantics can realise any set of models, the subclass of bipolar ADFs turns out to be less expressive.
This is shown using the next result, which allows us to decide realisability of a given model set \mbox{$X\subseteq 2^A$} in non-deterministic polynomial time.
We assume that the size of the input is in the order of $\cardinality{2^A}$, that is, the input set $X$ is represented directly.
The decision procedure then basically uses the construction of \thref{thm:realise:model} and an additional encoding of bipolarity to define a reduction to the satisfiability problem in propositional logic.
\begin{theorem}
  \label{thm:bipolar-realisability}
  Let \mbox{$X\subseteq 2^A$} be a set of sets.
  It is decidable in non-deterministic polynomial time whether there exists a bipolar ADF $D$ with \mbox{$\su(D)=X$}.
  \begin{proof}
    We construct a propositional formula $\phi_X$ that is satisfiable if and only if $X$ is bipolarly realisable.
    The propositional signature we use is the following:
    For each \mbox{$a\in A$} and \mbox{$M\subseteq A$}, there is a propositional variable $\pin{M}{a}$ that expresses whether \mbox{$C_a(M)=\tvt$}.
    This allows to encode all possible acceptance conditions for the statements in $A$.
    To enforce bipolarity, we use additional variables to model supporting and attacking links:
    for all \mbox{$a,b\in A$}, there is a variable $\psup{a}{b}$ saying that $a$ supports $b$, and a variable $\patt{a}{b}$ saying that $a$ attacks $b$.
    So the vocabulary of $\phi_X$ is given by
    \begin{displaymath}
      P = \set{ \pin{M}{a}, \psup{a}{b}, \patt{a}{b} \guard M\subseteq A, a\in A, b\in A }    
    \end{displaymath}
    To guarantee the desired set of models, we constrain the acceptance conditions as dictated by $X$:
    For any desired set $M$ and statement $a$, the containment of $a$ in $M$ must correspond exactly to whether \mbox{$C_a(M)=\tvt$};
    this is encoded in $\phi_X^{\in}$.
    Conversely, for any undesired set $M$ and statement $a$, there must not be any such correspondence, which $\phi_X^{\notin}$ expresses.
    \begin{align*}
      \phi_X^{\in} &= \bigwedge_{M\in X}\left( \bigwedge_{a\in M}\pin{M}{a} \wedge \bigwedge_{a\in A\setminus M}\neg\pin{M}{a} \right) \\
      \phi_X^{\notin} &= \bigwedge_{M\subseteq A, M\notin X}\left( \bigvee_{a\in M}\neg\pin{M}{a} \vee \bigvee_{a\in A\setminus M}\pin{M}{a} \right)
    \end{align*}
    To enforce bipolarity, we state that each link must be supporting or attacking.
    To model the meaning of support and attack, we encode all ground instances of their definitions.
    \begin{align*}
      \phi_{\mathit{bipolar}} &= \bigwedge_{a,b\in A}\left( \left(\psup{a}{b} \lor \patt{a}{b}\right) \land \phi_{\bsup}^{a,b} \land \phi_{\batt}^{a,b} \right) \\
      \phi_\bsup^{a,b} &= \psup{a}{b} \limplies \bigwedge_{M\subseteq A} \left( \pin{M}{b} \limplies \pin{M\cup\set{a}}{b} \right) \\ 
      \phi_\batt^{a,b} &= \patt{a}{b} \limplies \bigwedge_{M\subseteq A} \left( \pin{M\cup\set{a}}{b} \limplies \pin{M}{b} \right) 
    \end{align*}
    The overall formula is given by \mbox{$\phi_X = \phi_X^{\in} \land \phi_X^{\notin} \land \phi_{\mathit{bipolar}}$}.
    The rest of the proof -- showing that $X$ is bipolarly realisable if and only if $\phi_X$ is satisfiable -- is delegated to \Cref{thm:bipolar-claim} in the Appendix.
  \end{proof}
\end{theorem}

Remarkably, the decision procedure does not only give an answer, but in the case of a positive answer we can read off the BADF realisation from the satisfying evaluation of the constructed formula.
We illustrate the construction with an example that will subsequently be used to show that general ADFs are strictly more expressive than bipolar ADFs.

\begin{example}
  \label{exm:bipolar-unrealisable}
  Consider \mbox{$A=\set{x,y,z}$} and this model set:
  $$ X_1 = \set{ \emptyset, \set{x,y}, \set{x,z}, \set{y,z} } $$
  The construction of \thref{thm:bipolar-realisability} yields these formulas\/:
  \def\somespace{\quad\ \,}
  \begin{align*}
    \phi_{X_1}^{\in} &= \neg \pin{\emptyset}{x} \land \neg \pin{\emptyset}{y} \land \neg \pin{\emptyset}{z} \ \land \\
    &\somespace\pin{\set{x, y}}{x} \land \pin{\set{x, y}}{y} \land \neg \pin{\set{x, y}}{z}\ \land \\
    &\somespace\pin{\set{x, z}}{x} \land \neg\pin{\set{x, z}}{y} \land \pin{\set{x, z}}{z}\ \land \\
    &\somespace\neg \pin{\set{y, z}}{x} \land \pin{\set{y, z}}{y} \land \pin{\set{y, z}}{z} \\
    \phi_{X_1}^{\notin} &= ( \neg \pin{\set{x}}{x} \lor \pin{\set{x}}{y} \lor \pin{\set{x}}{z} )\ \land \\
    & \somespace ( \pin{\set{y}}{x} \lor \neg \pin{\set{y}}{y} \lor \pin{\set{y}}{z} )\ \land \\
    & \somespace ( \pin{\set{z}}{x} \lor \pin{\set{z}}{y} \lor \neg \pin{\set{z}}{z} )\ \land \\
    & \somespace ( \neg \pin{\set{x, y, z}}{x} \lor \neg \pin{\set{x, y, z}}{y} \lor \neg \pin{\set{x, y, z}}{z} )
  \end{align*}
  The remaining formulas about bipolarity are independent of $X_1$, we do not show them here.
  We have implemented the translation of \thref{thm:bipolar-realisability} and used the solver clasp~\citep{potassco} to verify that $\phi_{X_1}$ is unsatisfiable.
\end{example}

A manual proof of bipolar non-realisability of $X_1$ seems to amount to a laborious case distinction that explores the mutual incompatibility of the disjunctions in $\phi_{X_1}^{\notin}$ and bipolarity, a task that is better left to machines.
Together with the straightforward statement of fact that $X_1$ can be realised by a non-bipolar ADF, the example leads to the next result.

\begin{theorem}
  \label{thm:badfsu-v-adfsu}
  $\BADFsu \elt \ADFsu$
  \begin{proof}
    The model set from \Cref{exm:bipolar-unrealisable} is realisable under model semantics by ADF $D_{X_1}$ with acceptance conditions
    \begin{displaymath}
      \varphi_x = (y\lniff z),\quad \varphi_y = (x\lniff z),\quad \varphi_z = (x\lniff y)
    \end{displaymath}
    where ``$\lniff$'' denotes exclusive disjunction XOR.
    However, there is no bipolar ADF realising the model set $X_1$, as is witnessed by unsatisfiability of $\phi_{X_1}$ and \thref{thm:bipolar-realisability}.
  \end{proof}
\end{theorem}
Clearly ADF $D_{X_1}$ is not bipolar since in all acceptance formulas, all statements are neither supporting nor attacking.
It is not the only realisation, some alternatives are given by
\begin{align*}
  D_{X_1}':\quad&\varphi_x = (y\lniff z),\qquad \varphi_y = y,\qquad \varphi_z = z \\
  D_{X_1}'':\quad&\varphi_x = x,\qquad \varphi_y = (x\lniff z),\qquad \varphi_z = z \\
  D_{X_1}''':\quad&\varphi_x = x,\qquad \varphi_y = y,\qquad \varphi_z = (x\lniff y)
\end{align*}
This shows that we cannot necessarily use the model set $X_1$ to determine a \emph{single} reason for bipolar non-realisability, that is, a \emph{single} link \mbox{$(b,a)$} that is neither supporting nor attacking in \emph{all} realisations.
Rather, the culprit(s) might be different in each realisation, and to show bipolar non-realisability, we have to prove that \emph{for all} realisations, there necessarily \emph{exists some} reason for non-bipolarity.
And the number of different ADF realisations of a given model set $X$ can be considerable, as our next result shows.

\begin{proposition}
  \label{thm:adf-realisable-number}
  Let \mbox{$\cardinality{A}=n$}, \mbox{$X\subseteq 2^A$} with \mbox{$\cardinality{2^A\setminus X}=m$}.
  The number of distinct ADFs $D$ with \mbox{$\su(D)=X$} is
  \begin{center}
    $\displaystyle r(n,m) = \left(2^n-1\right)^m$
  \end{center}%
  \begin{proof}
    We have to count the number of distinct models of the formula
    \mbox{$\phi_X'=\phi_X^{\in}\land\phi_X^{\notin}$} from the proof of \thref{thm:bipolar-realisability}.
    We first observe that for each \mbox{$a\in A$} and \mbox{$M\subseteq A$}, the propositional variable $\pin{M}{a}$ occurs exactly once in $\phi_X'$.
    Formula $\phi_X^{\in}$ is a conjunction of literals and does not contribute to combinatorial explosion.
    Formula $\phi_X^{\notin}$ contains $m$ conjuncts.
    Each of the conjuncts is a disjunction of $n$ distinct literals.
    There are $2^n-1$ ways to satisfy such a disjunction.
    The claim now follows since for each of $m$ conjuncts, we can choose one of $2^n-1$ different ways to satisfy it.
  \end{proof}
\end{proposition}

So the main contributing factor is the number $m$ of interpretations that are excluded from the desired model set~$X$.
For \Cref{exm:bipolar-unrealisable}, for instance, there are \mbox{$(2^3-1)^{4}=7^4=2401$} ADFs with the model set $X_1$.
According to \thref{thm:badfsu-v-adfsu}, none of them is bipolar.
Obviously, the maximal number of realisations is achieved by \mbox{$X=\emptyset$} whence 
\mbox{$r(n,2^n)=(2^n-1)^{2^n}$}.
On the other hand, the model set \mbox{$X=2^A$} has exactly one realisation, \mbox{$r(n,0)=1$}.


It is comparably easy to show that BADF models are strictly more expressive than AFs, since sets of supported models of bipolar ADFs do not have the antichain property.

\begin{proposition}
  \label{thm:af-vs-badf}
  $\AF \elt \BADFsu$
  \begin{proof}
    Consider the vocabulary \mbox{$A=\set{a}$} and the BADF
    \mbox{$D=(A,\set{(a,a)}, \set{\varphi_a})$} with
    \mbox{$\varphi_a=a$}.
    It is straightforward to check that its model set is \mbox{$\su(D)=\set{\emptyset,\set{a}}$}.
    Since model sets of AFs under stable extension semantics satisfy the antichain property, there is no equivalent AF over $A$.
  \end{proof}
\end{proposition}

This yields the following overall relationships:
$$\AF \elt \BADFsu \elt \ADFsu \eeq \LPsu \eeq \PL$$

\subsection{Stable semantics}
\label{sec:expressiveness:stable}

As before, we recall the current state of knowledge:
\begin{gather*}
  \AF \eleq \BADFst \eleq \ADFst \elt \PL
  \text{ and }
  \AF \eleq \LPst \elt \PL
\end{gather*}

We first show that BADFs are strictly more expressive than AFs.
\begin{proposition}
  \label{thm:badfst-vs-af}
  $\AF \elt \BADFst$
  \begin{proof}
    Consider the BADF from \thref{thm:af-vs-badf}, where the acceptance formula of the single statement $a$ is given by \mbox{$\varphi_a = a$}. 
    Its only stable model is $\emptyset$.
    However there is no AF with a single argument with the same set of stable extensions:
    the only candidates are \mbox{$(\set{a},\emptyset)$} and \mbox{$(\set{a},\set{(a,a)})$};
    their respective stable-extension sets are $\set{\set{a}}$ and $\emptyset$.
  \end{proof}
\end{proposition}

Even if we discount for this special case of realising the empty stable extension, there are non-trivial extension-sets that AFs cannot realise.
\begin{example}[{\normalfont\citep{dunne14characteristics}}]
  \label{exm:unrealisable-AF}
  Consider the model set \mbox{$X_2=\set{ \set{x,y}, \set{x,z}, \set{y,z} }$}.
  \citet{dunne14characteristics} proved that $X_2$ is not realisable with stable AF semantics.
  Intuitively, the argument is as follows:
  Since $x$ and $y$ occur in an extension together, there can be no attack between them.
  The same holds for the pairs \mbox{$x,z$} and \mbox{$y,z$}.
  But then the set \mbox{$\set{x,y,z}$} is conflict-free and thus there must be a stable extension containing all three arguments, which is not allowed by $X_2$.
  The reason is AFs' restriction to individual attack, as set attack (also called joint or collective attack) suffices to realise $X_2$ with 
  BADF $D$ under stable model semantics:
  \begin{gather*}
    \varphi_x = \neg y \lor \neg z,
    \qquad
    \varphi_y = \neg x \lor \neg z,
    \qquad
    \varphi_z = \neg x \lor \neg y
  \end{gather*}
  Let us exemplarily show that \mbox{$M=\set{x,y}$} is a stable model (the other cases are completely symmetric):
  The reduct $D^M$ is characterised by the two acceptance formulas
  \mbox{$\varphi_x=\neg y\lor \neg\tvf$} and \mbox{$\varphi_y=\neg x\lor \neg \tvf$}.
  We then easily find that \mbox{$\Gamma_{D^M}(\emptyset, \emptyset) = (M,\emptyset) = \Gamma_{D^M}(M, \emptyset)$}.
\end{example}

The construction from the previous example model set comes from logic programming~\citep{eiter13modelbased} and can be generalised to realise any non-empty model set satisfying the antichain property.

\begin{definition}
  \label{def:canonical-badf}
  Let \mbox{$X\subseteq 2^A$}.
  Define the following BADF \mbox{$D^\st_X=(A,L,C)$} where $C_a$ for \mbox{$a\in A$} is given by
  \begin{gather*}
    \varphi_a = \bigvee_{M\in X,a\in M}\left(\bigwedge_{b\in A\setminus M}\neg b\right)
  \end{gather*}
  and thus \mbox{$L = \set{ (b,a) \guard M\in X,a\in M, b\in A\setminus M }$}.
\end{definition}

We next show that the construction indeed works.
\begin{theorem}
  \label{thm:badfst-realisable}
  Let $X$ with \mbox{$\emptyset\neq X\subseteq 2^A$} be a $\subseteq$-antichain.
  We find that \mbox{$\st(D^\st_X)=X$}.
  \begin{proof}
    Let \mbox{$M\subseteq A$}.
    \begin{description}
    \item[\normalfont ``$\subseteq$'':]
      Let \mbox{$M\notin X$}.
      We show that \mbox{$M\notin\su(D^\st_X)\supseteq\st(D^\st_X)$}.
      \begin{enumerate}
      \item There is an \mbox{$N\in X$} with \mbox{$M\subsetneq N$}.
        Then there is an \mbox{$a\in N\setminus M$}.
        Consider its acceptance formula $\varphi_a$.
        Since \mbox{$a\in N$} and \mbox{$N\in X$}, the formula $\varphi_a$ has a disjunct
        \mbox{$\psi_{a,N}=\bigwedge_{b\in A\setminus N}\neg b$}.
        Now \mbox{$M\subseteq N$} implies \mbox{$A\setminus N \subseteq A\setminus M$} and $M$ is a model for $\psi_{a,N}$.
        Thus $M$ is a model for $\varphi_a$ although \mbox{$a\notin M$}, hence \mbox{$M\notin\su(D^\st_X)$}.
      \item For all \mbox{$N\in X$}, we have \mbox{$M\not\subseteq N$}.
        Obviously \mbox{$M\neq\emptyset$} since \mbox{$X\neq\emptyset$}.
        Let \mbox{$a\in M$}. 
        For each \mbox{$N\in X$} with \mbox{$a\in N$}, the acceptance formula $\varphi_a$ contains a disjunct \mbox{$\psi_{a,N}=\bigwedge_{b\in A\setminus N}\neg b$}.
        By assumption, for each \mbox{$N\in X$} there is a \mbox{$b_N\in M\setminus N$}.
        Clearly \mbox{$b_N\in A\setminus N$} and $b_N$ is evaluated to true by $M$.
        Hence for each \mbox{$N\in X$} with \mbox{$a\in N$}, the disjunct $\psi_{a,N}$ is evaluated to false by $M$.
        Thus $\varphi_a$ is false under $M$ and \mbox{$M\notin\su(D^\st_X)$}.
      \end{enumerate}        
    \item[\normalfont ``$\supseteq$'':] 
      Let \mbox{$M\in X$}.
      We first show that $M$ is a model of $D^\st_X$,
      that is: for all \mbox{$a\in A$}, \mbox{$a\in M$} iff $M$ is a model for $\varphi_a$.
      \begin{enumerate}
      \item Let \mbox{$a\in M$}.
        By construction, we have that $\varphi_a$ in $D^\st_X$ contains a disjunct of the form \mbox{$\psi_{a,M}=\bigwedge_{b\in A\setminus M}\neg b$}.
        According to the interpretation $M$, all such \mbox{$b\in A\setminus M$} are false and thus $\psi_{a,M}$ is true whence $\varphi_a$ is true. 
      \item Let \mbox{$a\in A\setminus M$} and consider its acceptance formula $\varphi_a$.
        Assume to the contrary that $M$ is a model for $\varphi_a$.
        Then there is some \mbox{$N\in X$} with \mbox{$a\in N$} such that $M$ is a model for
        \mbox{$\psi_{a,N}=\bigwedge_{b\in A\setminus N}\neg b$}, that is,
        \mbox{$A\setminus N\subseteq A\setminus M$}.
        Hence \mbox{$M\subseteq N$} and $X$ is not a $\subseteq$-antichain.
        Contradiction.
        Thus $M$ is no model for $\varphi_a$.
      \end{enumerate}
      Now consider the reduct $D^M$ of $D^\st_X$ with respect to $M$.
      There, $\varphi_a^M$ contains the disjunct \mbox{$\psi_{a,M}^M=\psi_{a,M}[b/\tvf : b\notin M]$} where all \mbox{$b\in A\setminus M$} have been replaced by false, whence \mbox{$\psi_{a,M}^M = \neg\tvf \land \ldots \land \neg\tvf$} and $\varphi_a^M$ is equivalent to true.
      Thus each \mbox{$a\in M$} is true in the least fixpoint of $\Gamma_{D^M}$ and thus \mbox{$M\in \st(D^\st_X)$}.
    \end{description}
  \end{proof}
\end{theorem}
The restriction to non-empty model sets is immaterial, since we can use the construction of \thref{thm:realise:model} to realise the empty model set.

Since the stable model semantics for both ADFs and normal logic programs have the antichain property, the following is clear.
\begin{corollary}
  \mbox{$\ADFst \eleq \BADFst$} and \mbox{$\LPst \eleq \BADFst$}
\end{corollary}

For the family of stable semantics, this leads to the following overall expressiveness relationships:
$$\AF \elt \BADFst \eeq \ADFst \eeq \LPst \elt \PL$$

\subsection{Supported vs.\ stable semantics}
\label{sec:expressiveness:supported-v-stable}

Now we put the supported and stable pictures together.
From the proof of \thref{thm:badfst-realisable}, we can read off that for the canonical realisation $D^\st_X$ of an antichain $X$, the supported and stable semantics coincide, that is, \mbox{$\su(D^\st_X)=\st(D^\st_X)=X$}.
With this observation, also bipolar ADFs under the supported semantics can realise any antichain, and we have this:

\begin{proposition}
  $\BADFst \eleq \BADFsu$
\end{proposition}

As we have seen in \thref{thm:af-vs-badf}, there are bipolar ADFs with supported-model sets that are not antichains.
Thus we get the following result.

\begin{corollary}
  $\BADFst \elt \BADFsu$  
\end{corollary}

This result allows us to close the last gap and put together the big picture in \Cref{fig:hierarchy} below.

\begin{figure}[h]
  \centering
  \begin{tikzpicture}
    \node (af) at (0,0) {$\AF$};
    \node (st) at (0,1) {$\BADFst \eeq \ADFst \eeq \LPst$};
    \node (sb) at (0,2) {$\BADFsu$};
    \node (pl) at (0,3) {$\ADFsu \eeq \LPsu \eeq \PL$};

    \path[-] (af) edge (st);
    \path[-] (st) edge (sb);
    \path[-] (sb) edge (pl);
  \end{tikzpicture}
  \caption{The expressiveness hierarchy. 
    Expressiveness strictly increases from bottom to top. 
    $L^\sigma$ denotes language $L$ under semantics $\sigma$, where ``$\su$'' is the supported and ``$\st$'' the stable model semantics;
    languages are among 
    AFs (argumentation frameworks),
    ADFs (abstract dialectical frameworks),
    BADFs (bipolar ADFs),
    LPs (normal logic programs) and
    PL (propositional logic).
  }
    \label{fig:hierarchy}
\end{figure}

\section{Discussion}
\label{sec:discussion}

We compared the expressiveness of abstract argumentation frameworks, abstract dialectical frameworks, normal logic programs and propositional logic.
We showed that expressiveness under different semantics varies for the formalisms and obtained a neat expressiveness hierarchy.
These results inform us about the capabilities of these languages to encode sets of two-valued interpretations, and help us decide which languages to use for specific applications.

For instance, if we wish to encode arbitrary model sets, for example when using model-based revision, then ADFs and logic programs under supported semantics are a good choice.
If we are happy with the restricted class of model sets having the antichain property, then we would be ill-advised to use general ADFs under stable model semantics with their \SigmaP{2}-hard stable model existence problem;
to realise an antichain, it suffices to use bipolar ADFs or normal logic programs, where stable model existence is in \NP.

There is much potential for further work.
First of all, for results on non-realisability, it would be better to have necessary conditions than having to use a non-deterministic decision procedure.
For this, we need to obtain general criteria that all model sets of a given formalism must obey, given the formalism is not universally expressive.
This is non-trivial in general, and for AFs it constitutes a major open problem~\citep{dunne14characteristics,baumann14compact}.
Likewise, we sometimes used semantical realisations instead of syntactic ones;
for example, to show universal realisability of ADFs under supported models we started out with model sets.
It is an interesting question whether a realising ADF can be constructed from a given propositional formula without computing the models of the formula first.
Second, there are further semantics for abstract dialectical frameworks whose expressiveness could be studied;
\citet{dunne14characteristics} already analyse many of them for argumentation frameworks.
This work is thus only a start and the same can be done for the remaining semantics, for example admissible, complete, preferred and others, which are all defined for AFs, (B)ADFs and LPs~\citep{strass13approximating,brewka13adfs}.
Third, there are further formalisms in abstract argumentation~\citep{brewka13survey} whose expressiveness is by and large unexplored to the best of our knowledge.
Fourth, the requirement that realisations may only use a fixed vocabulary without any additional symbols is quite restrictive.
Intuitively, it should be allowed to add a reasonable number of additional atoms, for example a constant number or one that is linear in the original vocabulary.
Finally, our study only considered \emph{if} a language can express a model set, but not \emph{to what cost} in terms of representation size.
So the natural next step is to consider the succinctness of formalisms, ``How large is the smallest knowledge base expressing a given model set?''~\citep{gogic95comparative}.
A landmark result in this direction has been obtained by \citet{lifschitz-razborov06why}, who have shown that logic programs (with respect to two-valued stable models) are exponentially more succinct than propositional logic.
That is, there are logic programs whose respective sets of stable models cannot be expressed by a propositional formula whose size is at most polynomial in the size of the logic program, unless a certain widely believed assumption of complexity theory is false.
With the results of the present paper, we have laid the groundwork for a similar analysis of the other knowledge representation languages considered here, perhaps working towards a ``map'' of these languages in the sense of \citeauthor{darwiche-marquis02aknowledgecompilationmap}' knowledge compilation map~[\citeyear{darwiche-marquis02aknowledgecompilationmap}].


\vspace*{-5mm}
\paragraph{Acknowledgements.} The author wishes to thank Stefan Woltran for providing a useful pointer to related work on realisability in logic programming, and Frank Loebe for several informative discussions.
This research was partially supported by DFG (project BR~1817/7-1).

\appendix

\section*{Appendix}

\begin{lemma}
  \label{thm:bipolar-claim}
  $X$ is bipolarly realisable if and only if the formula $\phi_X$ from \thref{thm:bipolar-realisability} is satisfiable.
  \begin{xproof}
    \begin{description}
    \item[\normalfont ``if'':]
      Let \mbox{$I\subseteq P$} be a model for $\phi_X$.
      For each \mbox{$a\in A$}, we define an acceptance condition as follows:
      for \mbox{$M\subseteq A$}, set \mbox{$C_a(M)=\tvt$} iff \mbox{$\pin{M}{a}\in I$}.
      It is easy to see that $\phi_{\mathit{bipolar}}$ guarantees that these acceptance conditions are all bipolar.
      The ADF is now given by \mbox{$D^\su_X = (A, A\times A, C)$}.
      It remains to show that any \mbox{$M\subseteq A$} is a model of $D^\su_X$ if and only if \mbox{$M\in X$}.
      \begin{description}
      \item[\normalfont ``if'':]
        Let \mbox{$M\in X$}.
        We have to show that $M$ is a model of $D^\su_X$.
        Consider any \mbox{$a\in A$}.
        \begin{enumerate}
        \item \mbox{$a\in M$}. Since $I$ is a model of $\phi^{\in}_X$, we have \mbox{$\pin{M}{a}\in I$} and thus by definition \mbox{$C_a(M)=\tvt$}.
        \item \mbox{$a\in A\setminus M$}. Since $I$ is a model of $\phi^{\in}_X$, we have \mbox{$\pin{M}{a}\notin I$} and thus by definition \mbox{$C_a(M)=\tvf$}.
        \end{enumerate}
      \item[\normalfont ``only if'':] 
        Let \mbox{$M\notin X$}.
        Since $I$ is a model of $\phi^{\notin}_X$, there is
        an \mbox{$a\in M$} such that \mbox{$C_a(M)=\tvf$} or
        an \mbox{$a\notin M$} such that \mbox{$C_a(M)=\tvt$}.
        In any case, $M$ is not a model of $D^\su_X$.
      \end{description}
    \item[\normalfont ``only if'':] 
      Let $D$ be a bipolar ADF with \mbox{$\su(D)=X$}.
      We use $D$ to define a model $I$ for $\phi_X$.
      First, for \mbox{$M\subseteq A$} and \mbox{$a\in A$}, set \mbox{$\pin{M}{a}\in I$} iff \mbox{$C_a(M)=\tvt$}.
      Since $D$ is bipolar, each link is supporting or attacking and for all \mbox{$a,b\in A$} we can find a valuation for $\psup{a}{b}$ and $\patt{a}{b}$.
      It remains to show that $I$ is a model for $\phi_X$.
      \begin{enumerate}
      \item $I$ is a model for $\phi_X^{\in}$: Since $D$ realises $X$, each \mbox{$M\in X$} is a model of $D$ and thus for all \mbox{$a\in A$} we have \mbox{$C_a(M)=\tvt$} iff \mbox{$a\in M$}.
      \item $I$ is a model for $\phi_X^{\notin}$: Since $D$ realises $X$, each \mbox{$M\subseteq A$} with \mbox{$M\notin X$} is not a model of $D$. Thus for each such $M$, there is an \mbox{$a\in A$} witnessing that $M$ is not a model of $D$:
        (1) \mbox{$a\in M$} and \mbox{$C_a(M)=\tvf$}, or 
        (2) \mbox{$a\notin M$} and \mbox{$C_a(M)=\tvt$}.
      \item $I$ is a model for $\phi_{\mathit{bipolar}}$:
        This is straightforward since $D$ is bipolar by assumption.
        \hfill\ensuremath{\Box}
      \end{enumerate}
    \end{description}
 \end{xproof}
\end{lemma}


\begin{thebibliography}{}

\bibitem[\protect\citeauthoryear{Baumann \bgroup et al.\egroup
  }{2014}]{baumann14compact}
Baumann, R.; Dvo\v{r}{\'a}k, W.; Linsbichler, T.; Strass, H.; and Woltran, S.
\newblock 2014.
\newblock Compact argumentation frameworks.
\newblock In Konieczny, S., and Tompits, H., eds., {\em Proceedings of the
  Fifteenth International Workshop on Non-Monotonic Reasoning (NMR)}.

\bibitem[\protect\citeauthoryear{Bidoit and
  Froidevaux}{1991}]{bidoit91negation}
Bidoit, N., and Froidevaux, C.
\newblock 1991.
\newblock Negation by default and unstratifiable logic programs.
\newblock {\em Theoretical Computer Science} 78(1):85--112.

\bibitem[\protect\citeauthoryear{Brewka and
  Woltran}{2010}]{brewka-woltran10adfs}
Brewka, G., and Woltran, S.
\newblock 2010.
\newblock {Abstract Dialectical Frameworks}.
\newblock In {\em Proceedings of the Twelfth International Conference on the
  Principles of Knowledge Representation and Reasoning (KR)},  102--111.

\bibitem[\protect\citeauthoryear{Brewka \bgroup et al.\egroup
  }{2013}]{brewka13adfs}
Brewka, G.; Ellmauthaler, S.; Strass, H.; Wallner, J.~P.; and Woltran, S.
\newblock 2013.
\newblock {Abstract Dialectical Frameworks Revisited}.
\newblock In {\em Proceedings of the Twenty-Third International Joint
  Conference on Artificial Intelligence (IJCAI)},  803--809.
\newblock IJCAI/AAAI.

\bibitem[\protect\citeauthoryear{Brewka, Dunne, and
  Woltran}{2011}]{brewka11relating}
Brewka, G.; Dunne, P.~E.; and Woltran, S.
\newblock 2011.
\newblock {Relating the Semantics of Abstract Dialectical Frameworks and
  Standard AFs}.
\newblock In {\em Proceedings of the Twenty-Second International Joint
  Conference on Artificial Intelligence (IJCAI)},  780--785.
\newblock IJCAI/AAAI.

\bibitem[\protect\citeauthoryear{Brewka, Polberg, and
  Woltran}{2013}]{brewka13survey}
Brewka, G.; Polberg, S.; and Woltran, S.
\newblock 2013.
\newblock Generalizations of {Dung} frameworks and their role in formal
  argumentation.
\newblock {\em IEEE Intelligent Systems} PP(99).
\newblock Special Issue on Representation and Reasoning. In press.

\bibitem[\protect\citeauthoryear{Clark}{1978}]{clark78negation}
Clark, K.~L.
\newblock 1978.
\newblock Negation as {F}ailure.
\newblock In Gallaire, H., and Minker, J., eds., {\em Logic and Data Bases},
  293--322.
\newblock Plenum Press.

\bibitem[\protect\citeauthoryear{Coste-Marquis \bgroup et al.\egroup
  }{2013}]{coste13revision}
Coste-Marquis, S.; Konieczny, S.; Mailly, J.-G.; and Marquis, P.
\newblock 2013.
\newblock On the revision of argumentation systems: Minimal change of arguments
  status.
\newblock {\em Proceedings of TAFA}.

\bibitem[\protect\citeauthoryear{Darwiche and
  Marquis}{2002}]{darwiche-marquis02aknowledgecompilationmap}
Darwiche, A., and Marquis, P.
\newblock 2002.
\newblock {A Knowledge Compilation Map}.
\newblock {\em Journal of Artificial Intelligence Research (JAIR)} 17:229--264.

\bibitem[\protect\citeauthoryear{Dimopoulos, Nebel, and
  Toni}{2002}]{dimopoulos02complexity}
Dimopoulos, Y.; Nebel, B.; and Toni, F.
\newblock 2002.
\newblock On the computational complexity of assumption-based argumentation for
  default reasoning.
\newblock {\em Artificial Intelligence} 141(1/2):57--78.

\bibitem[\protect\citeauthoryear{Dung}{1995}]{dung95acceptability}
Dung, P.~M.
\newblock 1995.
\newblock {On the Acceptability of Arguments and its Fundamental Role in
  Nonmonotonic Reasoning, Logic Programming and n-Person Games}.
\newblock {\em Artificial Intelligence} 77:321--358.

\bibitem[\protect\citeauthoryear{Dunne \bgroup et al.\egroup
  }{2014}]{dunne14characteristics}
Dunne, P.~E.; Dvořák, W.; Linsbichler, T.; and Woltran, S.
\newblock 2014.
\newblock Characteristics of {Multiple Viewpoints} in {Abstract Argumentation}.
\newblock In {\em Proceedings of the Fourteenth International Conference on the
  Principles of Knowledge Representation and Reasoning (KR)}.
\newblock To appear.

\bibitem[\protect\citeauthoryear{Eiter \bgroup et al.\egroup
  }{2013}]{eiter13modelbased}
Eiter, T.; Fink, M.; P{\"u}hrer, J.; Tompits, H.; and Woltran, S.
\newblock 2013.
\newblock Model-based recasting in answer-set programming.
\newblock {\em Journal of Applied Non-Classical Logics} 23(1--2):75--104.

\bibitem[\protect\citeauthoryear{Gebser \bgroup et al.\egroup
  }{2011}]{potassco}
Gebser, M.; Kaminski, R.; Kaufmann, B.; Ostrowski, M.; Schaub, T.; and
  Schneider, M.
\newblock 2011.
\newblock {Potassco: The Potsdam Answer Set Solving Collection}.
\newblock {\em AI Communications} 24(2):105--124.
\newblock Available at \texttt{http://potassco.sourceforge.net}.

\bibitem[\protect\citeauthoryear{Gelfond and
  Lifschitz}{1988}]{gelfond-lifschitz88thestablemodel}
Gelfond, M., and Lifschitz, V.
\newblock 1988.
\newblock {The Stable Model Semantics for Logic Programming}.
\newblock In {\em Proceedings of the International Conference on Logic
  Programming (ICLP)},  1070--1080.
\newblock The MIT Press.

\bibitem[\protect\citeauthoryear{Gogic \bgroup et al.\egroup
  }{1995}]{gogic95comparative}
Gogic, G.; Kautz, H.; Papadimitriou, C.; and Selman, B.
\newblock 1995.
\newblock The comparative linguistics of knowledge representation.
\newblock In {\em Proceedings of the Fourteenth International Joint Conference
  on Artificial Intelligence (IJCAI)},  862--869.
\newblock Morgan Kaufmann.

\bibitem[\protect\citeauthoryear{Lifschitz and
  Razborov}{2006}]{lifschitz-razborov06why}
Lifschitz, V., and Razborov, A.
\newblock 2006.
\newblock Why are there so many loop formulas?
\newblock {\em ACM Transactions on Computational Logic} 7(2):261--268.

\bibitem[\protect\citeauthoryear{Lin and Zhao}{2004}]{lin-zhao04assat}
Lin, F., and Zhao, Y.
\newblock 2004.
\newblock {ASSAT}: {C}omputing {A}nswer {S}ets of a {L}ogic {P}rogram by {SAT}
  {S}olvers.
\newblock {\em Artificial Intelligence} 157(1-2):115--137.

\bibitem[\protect\citeauthoryear{Marek and
  Truszczy{\'n}ski}{1991}]{marek91autoepistemic}
Marek, V.~W., and Truszczy{\'n}ski, M.
\newblock 1991.
\newblock Autoepistemic logic.
\newblock {\em Journal of the ACM} 38(3):587--618.

\bibitem[\protect\citeauthoryear{Osorio \bgroup et al.\egroup
  }{2005}]{osorio05inferring}
Osorio, M.; Zepeda, C.; Nieves, J.~C.; and Cort{\'e}s, U.
\newblock 2005.
\newblock Inferring acceptable arguments with answer set programming.
\newblock In {\em Proceedings of the Sixth Mexican International Conference on
  Computer Science (ENC)},  198--205.
\newblock IEEE Computer Society.

\bibitem[\protect\citeauthoryear{Strass and
  Wallner}{2014}]{strass-wallner14complexity}
Strass, H., and Wallner, J.~P.
\newblock 2014.
\newblock Analyzing the {Computational Complexity} of {Abstract Dialectical
  Frameworks} via {Approximation Fixpoint Theory}.
\newblock In {\em Proceedings of the Fourteenth International Conference on the
  Principles of Knowledge Representation and Reasoning (KR)}.
\newblock To appear.

\bibitem[\protect\citeauthoryear{Strass}{2013}]{strass13approximating}
Strass, H.
\newblock 2013.
\newblock Approximating operators and semantics for abstract dialectical
  frameworks.
\newblock {\em Artificial Intelligence} 205:39--70.

\end{thebibliography}
\end{document}